# Research on the Multiple Feature Fusion Image Retrieval Algorithm based on Texture Feature and Rough Set Theory

Xiaojie Shi[1, a], Yijun Shao[2, b]

[1]East China Normal University, School of Computer Science and Software Engineering;

[2]Department of Math&Econ, Dickinson College, USA.

**Abstract.** Recently, we have witnessed the explosive growth of images with complex information and content. In order to effectively and precisely retrieve desired images from a large-scale image database with low time-consuming, we propose the multiple feature fusion image retrieval algorithm based on the texture feature and rough set theory in this paper. In contrast to the conventional approaches that only use the single feature or standard, we fuse the different features with operation of normalization. The rough set theory will assist us to enhance the robustness of retrieval system when facing with incomplete data warehouse. To enhance the texture extraction paradigm, we use the wavelet Gabor function that holds better robustness. In addition, from the perspectives of the internal and external normalization, we re-organize extracted feature with the better combination. The numerical experiment has verified general feasibility of our methodology. We enhance the overall accuracy compared with the other state-of-the-art algorithms.

## 1. Introduction

With the rapid development of multimedia technology and the Internet, expanding image sources have been brought to the daily life, the high speed large capacity storage system provides a basic guarantee for the mass storage of target images, while various industries on the use of the image is more and more widely and the image information resource management and retrieval is increasingly important [1]. The image retrieval technology is being used in the areas of the medical image management, satellite remote sensing images analysis, the computer aided design and manufacturing, the geographic information system and criminal identification system of copyright [2]. According to the survey, the popular methodologies for image retrieval could be categorized as the four classes. (1) Content-based image retrieval (CBIR). (2) Semantic-based image retrieval. (3) Feedback-based image retrieval. (4) AI and Knowledge-based image retrieval.

1) Content-based image retrieval. CBIR uses the basic features of the color, shape and size existed independently in image system to achieve the retrieval goal. The analysis of the CBIR contains the epresentation of the index, index of the organization and the extraction of the indexes. Image index can be divided into manual extraction, the extraction of semi-automatic extraction and automatic extraction while this is largely depends on development of image processing technology. The representation of the image index can be divided into the keyword index, color characteristic index, shape characteristic index, index of texture feature, etc.

2) Semantic-based image retrieval. Unlike query based on the low-level physical features, semantic features query is based on text query which contains the natural language processing and traditional image retrieval technology. The primary goal of this technique is to minimize image visual features and semantic gap between the rich semantics and the gap is usualy eliminated by the feature transformation. Through jointing high-level semantics to the transformation of the low-level features, we could achieve image retrieval based on semantic without changing existing way of image feature databases and the matching patterns of the image [3-4].

3) Feedback-based image retrieval. The basic idea of the feedback based method is to allow the user to evaluate the results and marking, points out that the result of what the user wants to get from the query image, which is not related, then regard the user tags as the training sample feedback related information to the system to guide the next round of training and iteration. Image retrieval system based on the feedback can get more accurate search results and has great practical value while it needs the participation of the users as the supervised learning scenario.

4) AI and Knowledge-based image retrieval. Image retrieval system based on knowledge is based on knowledge in the field of artificial intelligence introduced into the field of image processing, the processing method based on image understanding, knowledge representation, machine learning, and combined with the prior knowledge of the experts and users to establish the image knowledge base and paradigm to realize intelligence of the image database retrieval. This technique is the combination of natural language processing, expert system and knowledge discovery.

Based on the methodologies with mentioned four aspects of techniques, in this paper, we propose the novel multiple feature fusion image retrieval algorithm based on the texture feature analysis and rough set theory. Besides enhancing the traditional feature based image retrieval algorithm, rough set theory is introduced and combined. This theory defines the knowledge as indistinguishable relationship of a cluster, this makes the knowledge has a clear mathematical sense, and can use mathematical method to process. It can analyze the truth hidden in the data without the need for any additional information which will enhance the robustness of traditional methods. Remainder of this paper is organized as follows. The section 2 describes the framework and detailed theoretical analysis on our proposed algorithm. The section 3 provides experiment verification and comparison analysis. Finally, the section 4 draws the conclusion of this paper.

## 2. Literature review

### 2.1 Color-based image retrieval

From the perspective of human vision, the color feature is one of the basic visual features of human perception and discrimination of different objects. Color, a kind of visual information attribute, less depends on the size, direction, and the angle of the picture. So it has strong robustness. Meantime, the computation of color feature is simple.

Color-based retrieval usually uses color histogram method by which any image can only give a picture of the histogram based on its color feature. But different images may have the same histogram. That is to say, histogram and image's relationship is one to many. So it is apparently inconsistent with human visual induction and has high rate error detection.

### 2.2 Texture-based image retrieval

Texture refers to the visual patterns that have properties of homogeneity that do not result from the presence of only a single color or intensity. It is an innate property of virtually all surfaces, including clouds, trees, bricks, hair, and fabric. It contains important information about the structural arrangement of surfaces and their relationship to the surrounding environment.

In the early 1970s, Haralick et al. proposed the co-occurrence matrix representation of texture features. This approach explored the gray level spatial dependence of texture. It first constructed a co-occurrence matrix based on the orientation and distance between image pixels and then extracted meaningful statistics from the matrix as the texture representation. However, its disadvantage is that the size of matrix is large and a wealth of redundant information in the matrix.

Tamura explored the texture representation from a different angle. They developed computational approximations to the visual texture properties found to be important in psychology studies. The six visual texture properties were coarseness, contrast, directionality, linelikeness, regularity, and roughness. One major distinction between the Tamura texture representation and the co-occurrence matrix representation is that all the texture properties in Tamura representation are visually meaningful, whereas some of the texture properties used in co-occurrence matrix representation may not be (for example, entropy). This characteristic makes the Tamura texture representation very attractive in image retrieval, as it can provide a more user-friendly interface. However, Tamura method is mainly used to deal with whole picture. It does not work well in the special area of the picture, for example focus of infection in the organ.

### 2.3 Shape-based image retrieval

In general, the shape representations can be divided into two categories, boundary-based and region-based. The former uses only the outer boundary of the shape while the latter uses the entire shape region. The most successful representatives for these two categories are Fourier descriptor and moment invariants.

The main idea of a Fourier descriptor is to use the Fourier transformed boundary as the shape feature. Some early work can be found in. To take into account the digitization noise in the image domain, Rui et al. proposed a modified Fourier descriptor which is both robust to noise and invariant to geometric transformations. But the extraction, description and matching of the shape are urging problems. Compared to the color-based and texture-based method, shape-method is difficult.

## 2.4 Semantic-based Image Retrieval

In order to solve handicaps of simple visualization characteristic, people proposed the semantic-based image retrieval. Compared physical characteristic of picture, semantic-based image retrieval depends on the text-based query, including natural language processing and traditional image retrieval technology.

Two aspects of semantic based image retrieval need to be solved: one is to provide high-level semantic description; the other one is to have the method of mapping the lower level visual features to high-level semantics. Because image visual feature and user visual data are inconsistent, so the visual features and the high-level semantic have huge gap. In view of this problem, Amoid divides the domain of knowledge into narrow and broad domain and Hermes used the similarity technology, directly converting the outdoor image to the natural language description of the scene.

Semantic-based image retrieval, it allows users to apply objective feeling to describe image. So it is able to improve efficiency and accuracy for users. But because of the existence of "semantic gap", it makes the semantic-based image retrieval faced with huge challenge.

## 3 The Proposed Algorithm

### 3.1 The Rough Set Theory

Theory of rough set theory is a new mathematical tool for dealing with the fuzzy and uncertain knowledge. Rough set theory is different from the basic theory of fuzzy set and the probability theory, it does not need to advance the number of a given certain characteristics or attributes, but directly from the set of a given problem description, the use of collection, lower approximation and describe the concept of uncertainty through indiscernibility relation and indiscernibility classes to determine the approximation of a given problem domain to find out the implicit knowledge and reveal the potential regularity. In short, the main ideas of the rough set theory is through the indiscernibility relation between attributes, on the premise of ability of classification unchanged, contracted through knowledge to export problem of the decision making and classification rules [5].

In the proposed image retrieval system, rough operation serves as the pre-processing step. For this goal, we introduce the basic concepts and definitions as preliminary foundation. In the knowledge based on rough set theory, the knowledge is regarded as a kind of the classifying observable objects. Suppose the $U$ represents the set constituted of the interested objects $U \notin \varnothing$ which is also called the domain of discourse. Any sub-set $X \subseteq U$ is the concept or category in $U$. Based on this, we have the knowledge base and the inseparable relationship that are defined as formula 1 and 2, respectively.

$$K = (U, R) \qquad (1)$$

$$[x]_{ind(P)} = \bigcap_{R \in P} [x]_R \qquad (2)$$

Where the $K$ represents the relationship system and $R$ is the family of equivalence relation on $U$. Rough set is based on set theory on the basis of the theory of classic. Classic set theory is only object for a category belongs to or does not belong to two kinds of relationship, but in rough set theory to the development of the third kind of relations: may be, or may not belong to. Based on this feature, the rough set has the basic features of information system, attributes reduction, attribute dependency and discernibility matrix.

Rough sets re-construct the objective world or the object abstraction as an information system, also known as attribute value system defined as formula 3.

$$S = (U, A, V, f) \tag{3}$$

Attribute reduction is one of the core content of rough set theory. As is known to all, the random sampling of data set, knowledge because of this randomness and deposited in the knowledge base is not equally important, and even some of the knowledge is redundant. Redundant knowledge on the one hand, is the waste of the resources, on the other hand, interferences with people to make concise decision. Attribute reduction is in keeping the database under the constant classification or decision ability while remove the irrelevant or unimportant knowledge. Through the attribute reduction, rough set theory can effectively find out the decisive factors, leading to remove data redundancy or secondary role factors to realize the data simplification and refinement. In addition, we define the attribute dependency in equation 4.

$$k = \gamma(D) = (pos_P(D))/(U) \tag{4}$$

### 3.2 The Texture Feature of Images.

Texture can be understood as can make the space in the form of a certain pattern, as a result of the change is a kind of doesn't depend on the color or brightness projection images in the visual characteristics of homogeneous phenomena. At present most of scholars believe that the classification of the texture analysis algorithm is the statistics method, structure method, model and space combined frequency domain (SCF) analysis method. Under the guidance of the prior research, we choose to use the Gabor wavelet to extract the features [6-7].

Due to the localization of frequency description needs a "window" of a fixed width in spatial domain, the frequency domain bandwidth is fixed on a fixed scale, so the frequency of the localized description is not enough completely suitable for character description. In order to optimize to detect the partial feature of different scales, it will require a different scale of the filter, filter rather than a fixed size. Therefore, we adopt the Gabor wavelet to assist extracting the targeted features. In the formula 5, we define the Gabor function.

$$g(x,y) = \left[\frac{1}{2\pi\sigma_x\sigma_y}\right]\exp\left[-\frac{1}{2}\left(\frac{x^2}{\sigma_x^2}+\frac{y^2}{\sigma_y^2}\right)+2\pi jWx\right] \tag{5}$$

As for the images, corresponding Gabor processed form could be defined as the equation 6, where the $I(x,y)$ is the image notation. Accordingly, the transform coefficient of the mean and the standard variance could be defined as the equations 7~8 under the prior condition of local texture spatial consistency.

$$GW_{m,n}(x,y) = \int I(x,y) g_{nm}^*(x-x_1, y-y_1) dx_1 dy_1 \tag{6}$$

$$\mu_{mn} = \iint |GW_{m,n}(x,y)| dxdy \tag{7}$$

$$\sigma_{mn} = \sqrt{\iint (GW_{m,n}(x,y) - \mu_{mn}) dxdy} \tag{8}$$

### 4.3 Multiple Feature Fusion based Retrieval Algorithm.

Integrated multiple feature fusion of the image retrieval technology is applicable only if the first normalization theory of image features are introduced briefly, because if we need to retrieve the image content, the system of image features generally need get combined with the characteristics of the multidimensional vector.

In the use of more features to retrieve, these must be the normalized transverse differences. That they have with the right values, but on behalf of the meaning of different, it can be directly compared. Theoretically, the vector normalization could be divided into the internal and external steps.

*Internal normalization.* Suppose the feature vector of the image is defined as $F = [f_1, f_2, \cdots, f_N]$, $M$ images in the database could construct the feature matrix $F = f_{ij}$ with the dimension of $M \times N$, the column of feature matrix denotes the feature sequence with $M$ dimension. The normalization steps could be achieved through the equation 9.

$$f_{i,j}^* = \frac{f_{i,j} - \mu_j}{\sigma_j} \tag{9}$$

*External normalization.* The external normalization refers to the steps of different external characteristic vector normalization processing. Comprehensive characteristics of each feature in the similarity comparison will be with same weight. Taking this way, after the processing of the image database, the similar distance measurement will be accurate. Specifically, in our framework, we use the equation 10 to get the normalized result.

$$D_i^* = \frac{1}{2}\left(1 + \frac{D_i - \mu_{Di}}{3\sigma_{Di}}\right) \tag{10}$$

After the normalization, we will then combine extracted features to retrieve images. As the performance evaluation reference, we adopt the Jensen Shannon Divergenc/Jeffrey Divergence distance (JSD) serve as the similarity computing standard. JSD, as the extended version of Kullback-Leibler Divergence holds better robustness. In the equation 11, we define the mentioned JSD.

$$d(\cdot) = \sum_{m=1}^{M}\left(H_m \log \frac{2H_m}{H_m + H_m'} + H_m' \log \frac{2H_m'}{H + H_m'}\right) \tag{11}$$

## 4 Experiment

### 4.1 Retrieval result

In this section, we present and show the simulation and verification on the effectiveness of the proposed algorithm. We firstly introduce the target database and the experimental environment and then we test the precision of the proposed method compared with other state-of-the-art algorithms. As the analysis basis, the numerical validation data is shown.

This section demonstrates the simulation on the image retrieval precision of the algorithms. In the table one and two, we illustrate the overall accuracy test with the different training image numbers and the different experimental image sizes. In the table three, we conduct comparison experiment on our proposed algorithm and the method1 [1], method2 [2] and method3 [3]. We could conclude from the result that the training image number and experiment image database will influence overall accuracy. With more training images, the algorithm will be better trained for retrieval because the general robustness and complexity of the algorithm will be well optimized. With more experiment images, the overall accuracy will be reduced due to the increase of the sample data sets that may influence the searching efficiency. Additionally, to show the visualized simulation of test, we demonstrate the sample retrieval result in figure one to help illustrate the performance visually.

TABLE I. OVERALL ACCURACY WITH DIFFERNET TRAINING IMAGES AND EXPERIMENT IMAGE SIZE SET ONE

| Images for Training | Experiment Image Size | Overall Accuracy | Experiment Image Size | Overall Accuracy |
|---|---|---|---|---|
| 5 | 1000 | 35.5% | 2000 | 33.2% |
| 15 | 1000 | 53.9% | 2000 | 51.2%. |
| 25 | 1000 | 68.2% | 2000 | 65.8% |
| 35 | 1000 | 73.9% | 2000 | 73.5% |
| 45 | 1000 | 86.2% | 2000 | 84.1% |
| 55 | 1000 | 88.3% | 2000 | 87.0% |
| 65 | 1000 | 90.8% | 2000 | 89.2% |
| 75 | 1000 | 92.7% | 2000 | 90.7% |

TABLE II. OVERALL ACCURACY WITH DIFFERENC TRAINING IMAGES AND EXPERIMENT IMAGE SIZE SET TWO

| Images for Training | Experiment Image Size | Overall Accuracy | Experiment Image Size | Overall Accuracy |
|---|---|---|---|---|
| 5 | 3000 | 31.5% | 4000 | 30.2% |

| Images for Training | *Experiment Image Size* | *Overall Accuracy* | *Experiment Image Size* | *Overall Accuracy* |
|---|---|---|---|---|
| 15 | 3000 | 49.3% | 4000 | 47.5% |
| 25 | 3000 | 62.8% | 4000 | 59.2% |
| 35 | 3000 | 71.5% | 4000 | 70.3% |
| 45 | 3000 | 83.5% | 4000 | 81.2% |
| 55 | 3000 | 86.7% | 4000 | 83.7% |
| 65 | 3000 | 88.5% | 4000 | 85.0% |
| 75 | 3000 | 89.8% | 4000 | 87.9% |

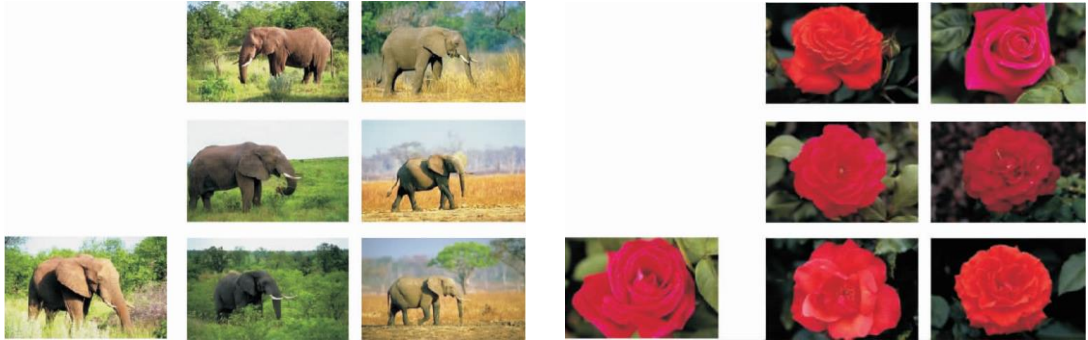

Fig. 1 The Sample Image One and the Retrieval Result

## 4.2 Segmentation result

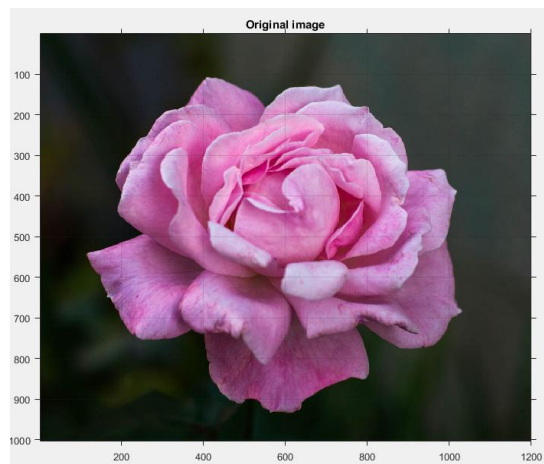

Original image 1

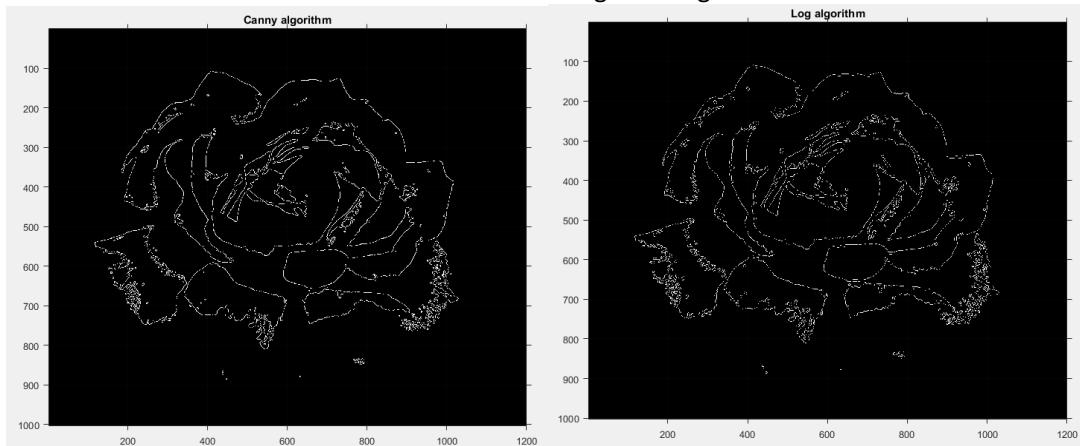

Canny Algorithm                                    Log Algorithm

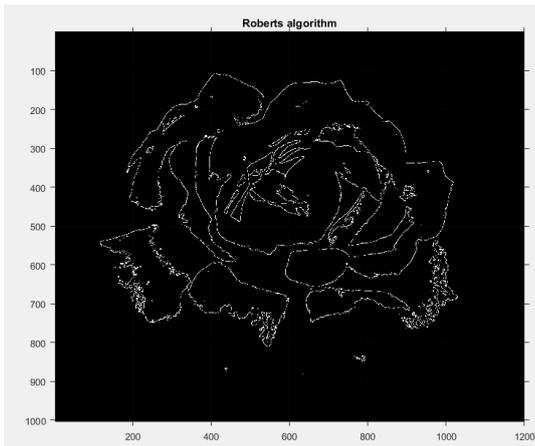
Roberts Algorithm

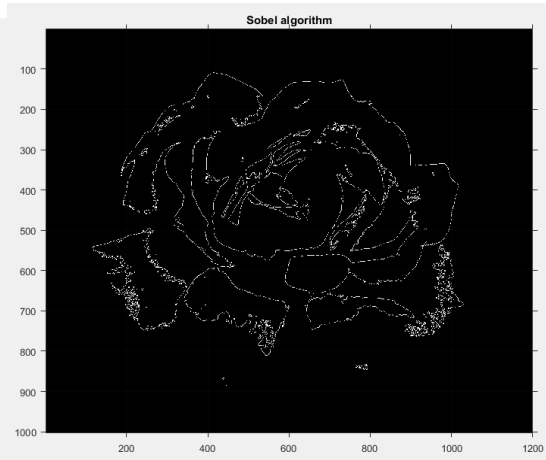
Sobel Algorithm

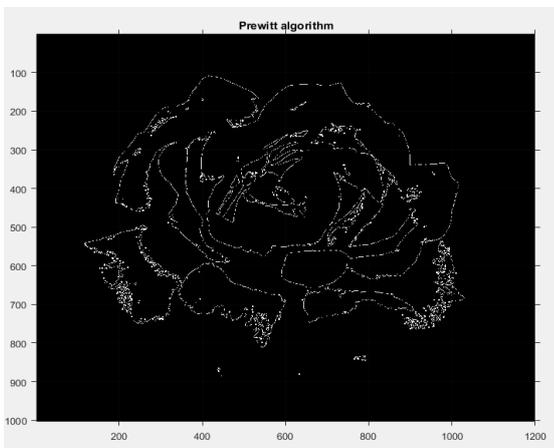
Prewitt Algorithm

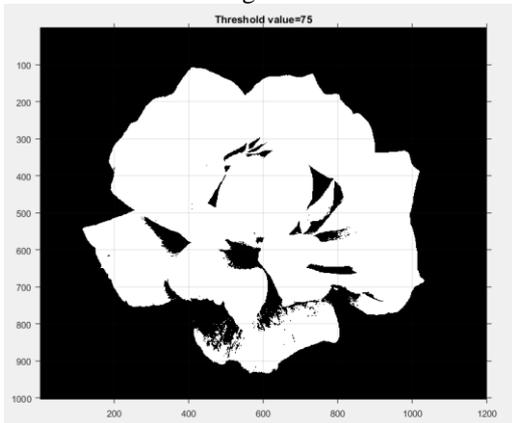
Threshold=75 Algorithm

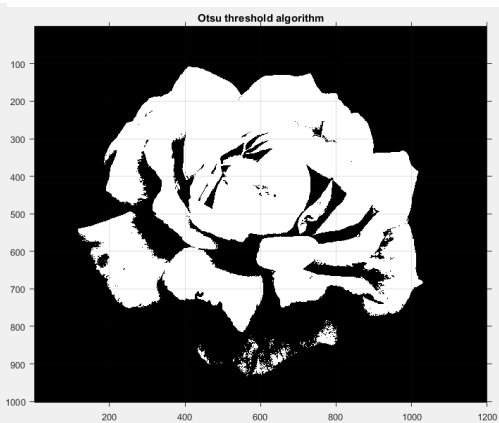
Ostu Algorithm

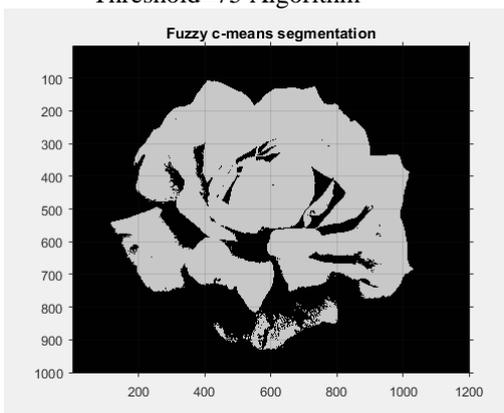
Fuzzy c-means

## 5 Summary

In this research paper, we study the problem of complex information retrieval (IR), and propose the multiple feature fusion image retrieval algorithm based on texture feature and rough set theory. In contrast to conventional approaches that only use the single feature or standard, we fuse the different features with operation of normalization. As implementation of the traditional content-based image retrieval algorithm in the aspect of dealing with incomplete data sets, we adopt the rough set analysis for optimization. To enhance the texture extraction paradigm, we use wavelet revised Gabor function that hold better robustness. In addition, from the perspectives of internal and external normalization, we re-organize feature with the better combination technique. Experimental results of proposed approach have shown significant improvement and modification in the overall retrieval accuracy visually and numerically. In the future, we will consider taking more feature extraction paradigms to achieve better combination of the information re-organization to eventually enhance the image searching accuracy.